\documentclass[11pt]{article}

\usepackage[final]{acl}

\usepackage{times}
\usepackage{latexsym}

\usepackage[T1]{fontenc}

\usepackage[utf8]{inputenc}

\usepackage{microtype}

\usepackage{inconsolata}

\usepackage{graphicx}

\usepackage{algorithm}
\usepackage{algorithmic}
\usepackage{enumitem}
\usepackage{amssymb}
\usepackage{amsmath}
\def\NN{\mathbb{N}}
\usepackage{tikz,pgfplots}
\usetikzlibrary{matrix}
\DeclareMathOperator*{\argmin}{arg\,min}

\title{Thinking Long, but Short:\\Stable Sequential Test-Time Scaling for Large Reasoning Models}

\author{Michael R. Metel, Yufei Cui, Boxing Chen, and Prasanna Parthasarathi\\
        Huawei Noah's Ark Lab\\
        \small{\textbf{Correspondence:} \href{mailto:michael.metel@huawei.com}{michael.metel@huawei.com}}}

\begin{document}
\maketitle
\begin{abstract}
Sequential test-time scaling is a promising training-free method to improve large reasoning model accuracy, but as currently implemented, significant limitations have been observed. Inducing models to think for longer can increase their accuracy, but as the length of reasoning is further extended, it has also been shown to result in accuracy degradation and model instability. This work presents a novel sequential test-time scaling method, {\it Min-Seek}, which improves model accuracy significantly over a wide range of induced thoughts, stabilizing the accuracy of sequential scaling, and removing the need for reasoning length fine-tuning. Beyond improving model accuracy over a variety of reasoning tasks, our method is inherently efficient, as only the KV pairs of one additional induced thought are kept in the KV cache during reasoning. With a custom KV cache which stores keys without position embeddings, by dynamically encoding them contiguously before each new generated thought, our method can continue to reason well beyond a model's maximum context length, and under mild conditions has linear computational complexity. 
\end{abstract}

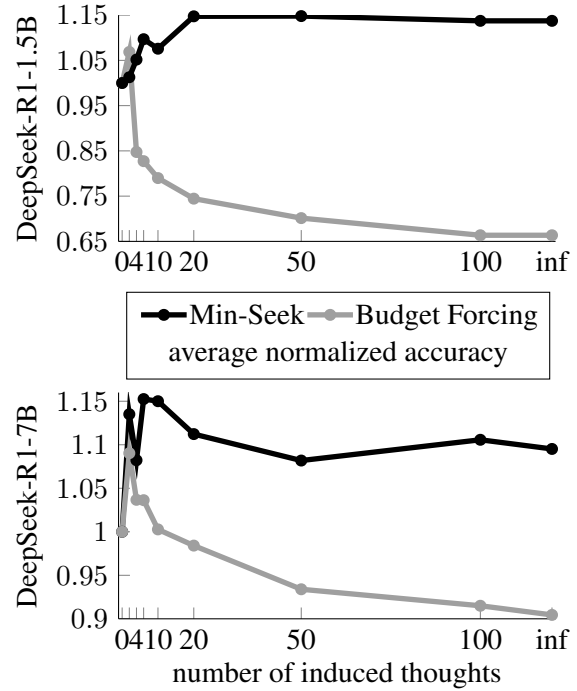
\begin{figure}[t]
\centering
\pgfplotsset{width=5.75cm,height=3cm,compat=1.3}
\begin{tikzpicture}
\pgfplotsset{scale only axis}
\begin{axis}[
axis y line*=left,axis x line*=bottom,
xmin=-1,xmax=121,ymin=0.65,ymax=1.15,
xtick=data,ytick={0.65,0.75,...,1.15},
xticklabels={0,,4,,10,20,50,100,inf},
ylabel=DeepSeek-R1-1.5B,
y label style={at={(axis description cs:-0.15,.5)},anchor=south}]
\addplot[draw=lightgray!50!gray,line width=2pt,mark=*,mark size=1.25pt,mark options={solid,fill=lightgray!50!gray}]
table[x=x,y=y]{add_thought_1B_norm.dat};\label{plot_five}	
\addplot[draw=black,line width=2pt,mark=*,mark size=1.25pt,mark options={solid,fill=black}]
table[x=x,y=y]{min_thought_1B_norm.dat};\label{plot_two}
\end{axis}
\matrix[matrix of nodes,anchor=west,xshift=0.1cm,yshift=-1.25cm,inner sep=0.2em,draw]
{\ref{plot_two}Min-Seek\ref{plot_five}Budget Forcing\\average normalized accuracy\\};
\end{tikzpicture}
\begin{tikzpicture}
\pgfplotsset{scale only axis}
\begin{axis}[
axis y line*=left,axis x line*=bottom,
xmin=-1,xmax=121,ymin=0.9,ymax=1.16,
xtick=data,ytick={0.9,0.95,...,1.15},
xticklabels={0,,4,,10,20,50,100,inf},
xlabel=number of induced thoughts,
ylabel=DeepSeek-R1-7B,
y label style={at={(axis description cs:-0.15,.5)},anchor=south},
x label style={at={(axis description cs:.5,-0.35)},anchor=south}]
\addplot[draw=black,line width=2pt,mark=*,mark size=1.25pt,mark options={fill=black}]
table[x=x,y=y]{min_thought_7B_norm.dat};\label{plot_four}
\addplot[draw=lightgray!50!gray,line width=2pt,mark=*,mark size=1.25pt,mark options={fill=lightgray!50!gray}]
table[x=x,y=y]{add_thought_7B_norm.dat};\label{plot_three}	
\end{axis}
\end{tikzpicture}
\caption{Average normalized accuracy over 5 reasoning tasks. An increasing number of thoughts (reconstruction cycles) were induced using Budget Forcing (grey) and our proposed Min-Seek (black) sequential test-time scaling method using DeepSeek-R1-Distill-Qwen-1.5B (above) and DeepSeek-R1-Distill-Qwen-7B (below).} \label{fig1}
\end{figure}

\section{Introduction}

With the development of large reasoning models (LRMs) such as OpenAI's o1 \cite{openai2024}, improved accuracy has been achieved by allocating more resources during test time, by training the model to generate longer outputs using chain-of-thought reasoning \cite{wei2022}. As a training-free method, forcing a model to reason for longer, referred to as test-time scaling, contains two main approaches: parallel and sequential. Parallel scaling generates multiple solutions to a problem, and among them, selects a final answer, such as by majority voting, where the most frequently generated answer is selected \cite{wang2023}. Sequential scaling instead generates an increasingly long reasoning chain before generating a single final answer. In this work the focus is on sequential test-time scaling, as it has the potential to build on past thoughts, exploring different solution approaches, verifying its reasoning at different stages, and progressively generating deeper and more complex thoughts through time. When the thinking tokens can be viewed by the end user, such as with DeepSeek-R1 \cite{deepseek2025}, a single coherent stream of reasoning is also more interpretable, and accommodating to follow-up prompting compared to parallel scaling, which can only offer several independent solutions.    

Although improved accuracy has been achieved, significant limitations of sequential test-time scaling have also been observed. Inducing an LRM to think for longer can improve its reasoning ability, but there tends to be a ``sweet spot", whereby over passing this unknown amount of thinking can destabilize the model. This work seeks to stabilize sequential scaling of LRMs, with the goal of being able to generate arbitrarily long reasoning chains, going even beyond a model's maximum context length, using only a training-free method. 

To stabilize sequential test-time scaling, we propose to only keep an evolving subset of past thoughts' KV pairs in the KV cache. When deciding on which past thoughts to keep, a natural approach would be to try to select the most ``important" past thoughts, e.g., containing key information or reasoning steps. Our method can be seen as focusing on the complementary objective of trying to exclude the KV pairs of past thoughts which are causing the loss in model stability. Based on significant evidence from past research and our own analysis, the proposed method, { \it Min-Seek}, is based on keeping the shortest past generated thoughts. The motivation is to avoid flawed reasoning paths, during which LRMs have been shown to reason for longer while attempting to recover from reasoning errors, whereas sound logic generally results in more direct and hence shorter reasoning. 
To summarize, we believe that forcing the model to attend to shorter thoughts decreases the probability of following flawed reasoning paths, resulting in more stable and accurate output. 

Min-Seek was tested on a variety of reasoning tasks, and showed significantly improved and more stable accuracy over a wide range of generated thoughts compared to when using {\it Budget Forcing} sequential scaling, as proposed and further tested in \cite{muennighoff2025,zeng2025}, respectively, where induced thoughts are repeatedly generated and attended to through time. By limiting the KV cache to only contain a fixed number of thoughts, our method enables the generation of an unbounded number of thoughts, and has linear computational complexity, under mild conditions. To bypass a model's maximum context length, our method stores keys in the KV cache without position embeddings, which are dynamically applied contiguously before generating each new thought. A modified KV cache is used which temporarily stores a copy of the KV cache with position embeddings for fast generation, while also updating the KV cache without position embeddings for future thought generation.
Updating the KV cache to contain the shortest length thoughts through time further simplifies our method, as it 
does not require the storage of all past thoughts, nor the use of any type of retrieval mechanism, with its 
next state being only dependent on its current state and the currently generated thought. 

We now summarize the two main limitations of sequential test-time scaling, and what our proposed method achieves.

\vspace{0.15cm}
\noindent{\bf Limitation 1}: There is a model and problem specific optimal reasoning length. Over-extending reasoning results in a break down in the attention mechanism, resulting in unstable, repetitive output, and a decrease in accuracy. 
\begin{itemize}
\item LRM output using Min-Seek is now stabilized through time, achieving significantly improved accuracy compared to standard model generation or when using Budget Forcing, over a wide range of reasoning lengths. 
\end{itemize}
{\bf Limitation 2}: The KV cache increases linearly with reasoning length, resulting in a theoretical quadratic computational complexity and a fixed generation length upper bound in practice given a model's maximum context length. 
\begin{itemize}
\item A custom KV cache containing only a small subset of past thoughts is used, enabling unconstrained lengths of reasoning with linear computational complexity, and significantly faster generation in practice compared to Budget Forcing.  
\end{itemize}

\noindent Section \ref{lit_rev} summarizes past work on test-time scaling. Section \ref{method} presents Min-Seek, including details about our custom KV cache and thought selection rule. Section \ref{experiments} contains experimental results, with the paper concluding in Section \ref{conclusion}.

\section{Background \& Related Work}
\label{lit_rev}

We first briefly describe the output of DeepSeek-R1 and its distilled variants. These models are trained and prompted \cite[Table 1]{deepseek2025} to generate output using the following structure,\\

\noindent\texttt{<think> reasoning process here </think>\\<answer> answer here </answer>},\\

\noindent making it straightforward to separate a model's reasoning from its answer.
Three important papers related to our work will now be discussed, which all observe the following two main findings. 
\begin{itemize} 
\item Accuracy initially increases with reasoning length, after which 
the model exhibits some level of instability resulting in repetitive output and accuracy degradation.
\item On average, shorter answers are more accurate than longer answers.
\end{itemize}


\noindent A simple implementation of sequential test-time scaling, Budget Forcing, was proposed in \cite{muennighoff2025}. Whenever the model stops its reasoning chain with a {\tt <\textbackslash think>} token, it is replaced with the text ``Wait" to induce the model to continue thinking. Following \cite[Section 3.2]{marjanovic2025}, this thought can be more precisely referred to as a {\it reconstruction cycle}, with the model reconsidering its past assumptions and solutions, and possibly giving a new solution. Although this method was shown to increase accuracy for a limited number of additional thinking rounds (2, 4, and 6 new generated thoughts), it was found that doing this too often can result in the model becoming unstable with its output looping repetitively \cite[Section 4.2]{muennighoff2025}, which was also observed in our experiments. By performing rejection sampling, an inverse scaling trend was also observed, where on average the model's accuracy decreased as the number of thinking tokens increased. It is hypothesized \cite[Section 5.2]{muennighoff2025} that shorter answers occur when the model starts from a correct reasoning path leading directly to the correct answer, whereas for longer solutions, the model generated flawed reasoning which it then struggles to recover from.


These findings are corroborated by \cite{marjanovic2025}, where their conclusions concerning reasoning length versus model accuracy was based on binning a number of sampled solutions together based on their number of thinking tokens. It was found that there is a problem-dependent optimal range for thinking length, and that going beyond this range results in substantially lower accuracy \cite[Section 4.3]{marjanovic2025}, which can even lead to the model becoming overwhelmed, resulting in the generation of long incoherent texts \cite[Section 5.4]{marjanovic2025}. It was also observed that on average, the amount of thinking done for correct answers is shorter than for incorrect answers \cite[Figure 4.3]{marjanovic2025}. The authors show two examples, exemplifying the hypothesis of \cite{muennighoff2025}, where the model starts from and gets trapped in a bad reasoning path, as well as an example where self-verification during the reasoning process fails, where the model initially had the correct answer, but then changes its mind, and proceeds to follow a longer incorrect reasoning path \cite[Figures C.2 \& C.3]{marjanovic2025}.

As was done in \cite{muennighoff2025}, LRMs were iteratively prompted in \cite{zeng2025} with the words ``Wait" or ``Alternatively", and it was found that this often led to accuracy degradation, and at best, models exhibiting an initial improvement followed by oscillatory behaviour \cite[Figure 4]{zeng2025}. The authors also found that the average reasoning length of correct solutions is consistently shorter than for incorrect solutions \cite[Figure 1]{zeng2025}. In order to understand why increased reasoning does not lead to improved accuracy, the authors compared long and short solutions, and found that all models increase their length of reasoning predominantly through self-revisions (reconstruction cycles), and in particular, the number of generated ``Wait" tokens increases linearly with sequence length \cite[Figure 3(b)]{zeng2025}. It can be concluded that a strong correlation between the length of reasoning, the number of self-revisions, and model performance exists, and that it is perhaps the increase in the number of self-revisions, specifically, which is the main cause for model accuracy degradation. Given the lackluster performance of sequential test-time scaling, the authors instead proposed a new parallel test-time scaling method, Shortest Majority Voting, which can be viewed as a parallel counterpart to our sequential Min-Seek method, which adds a new weighting based on reasoning length to the majority voting selection rule. 




\subsection{Test-time Scaling for LLMs}

There have also been many techniques proposed to improve the reasoning of Large Language Models (LLMs) through test-time scaling, including training-free methods using special prompting. Intermediate thoughts are generated, evaluated, and pruned within a tree structure in Tree of Thoughts \cite{yao2023}, using task and reasoning step dependent prompting. Similar to writing multiple drafts of a manuscript, Self-Refine \cite{madaan2023} iteratively generates an output, feedback on the output, which is then used to refine the previously-generated output using a single LLM. Focusing on LLMs as agents, Reflexion \cite{shinn2023} proposes a type of ``verbal" reinforcement learning, where feedback is inserted into the prompt based on the evaluation of past trials. 

In terms of methods requiring LLM training, InftyThink \cite{yan2025} considers iterative model generation to overcome the quadratic complexity and maximum context length of LLMs for long reasoning. Their proposed method fine-tunes an LLM to iteratively generate a new thought and a summary, which is then used with the original question as the next prompt, to allow the LLM to continue reasoning based on its summarized past output. For math reasoning with an LLM which has been trained to output solutions in a step-by-step format, a process-supervised reward model was found to work best to choose the best-of-N sampled solutions in \cite{lightman2024}. 

Since LRMs have already been trained to perform these types of complex reasoning techniques, these methods do not seem applicable for LRMs. 


\begin{figure*}[t]
	\includegraphics[width=2.05\columnwidth]{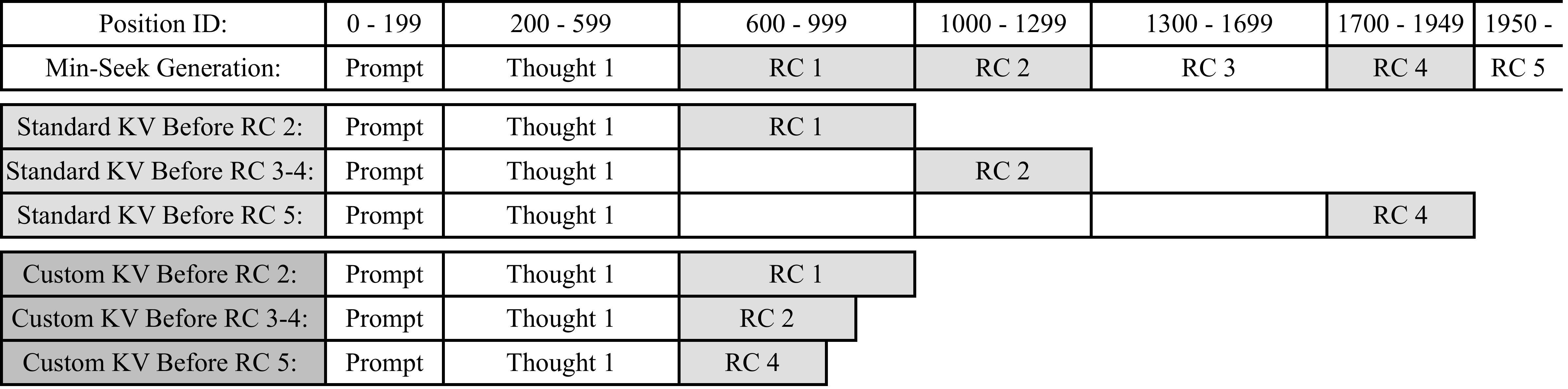}
	\caption{An example of Min-Seek while generating a fifth reconstruction cycle (RC 5). The first table shows the position IDs of the prompt and generated tokens, which are grouped together as Thought 1 and RC $i$. The second and third tables show the KV cache without and with our modification before generating RC 2-5, where the boxes, representing the keys and values, are placed according to the keys' position embeddings. The RC $i$ in grey are strictly shorter than all past reconstruction cycles.}
	\label{fig:experiments}
\end{figure*}

\section{Min-Seek Sequential Test-Time Scaling}
\label{method}

Based on the previously discussed research, improved accuracy is possible using sequential test-time scaling, but more work is needed to stabilize the method, as LRMs' accuracy tends to degrade as the number of self-revisions grows too large. Given that self-revisions may be contradictory, such as by exploring different solution approaches, as well as repetitive, as the number of self-revisions grows, we believe that the observation of the model becoming ``overwhelmed" is a breakdown in the self-attention mechanism, with the model being unable to focus on a single coherent chain of reasoning. At least empirically, LRM accuracy can be improved when only a limited number of reconstruction cycles are induced, so our first insight is to only allow the model to attend to a small number of past thoughts. More precisely, our method always keeps the KV cache of the prompt and the first generated  thought, denoted together as $PT_1$, but only the KV cache of a subset of induced reconstruction cycles through time, $\overline{RC}_r:=\{RC_i\}_{i\in I_r}$, where $RC_i$ is the $i^{\text{th}}$ generated reconstruction cycle, and $I_r$ indexes a chosen subset of the past $r$ generated reconstruction cycles. 

Given the autoregressive nature of LRMs, when removing the KV cache of a past reconstruction cycle, its information is not necessarily lost. The keys and values at layer $l\in\NN$ and position ID $t\in\NN$, $[K^l_t,V^l_t]$, are linear transformations of the hidden state, $X^l_t$, which is itself a function of 
$\{[K^j_s,V^j_s]\}^{j<l}_{s\leq t}$ (among other quantities), hence information contained in previous keys and values can be past on and contained in future KV pairs. In particular, we want the KV cache of future reconstruction cycles to improve upon past reconstruction cycles by retaining and building upon the information contained in their KV caches. As an abstract example, assume that idea $A\subset \text{(KV cache of) } RC_i$, idea $B\subset RC_j$, and $\{A,B\}\subset RC_{>\max\{i,j\}}$. If $\{A,B\}$ is all that is required from $RC_i$ and $RC_j$, then their KV caches can be discarded, as their useful information has already been synthesized into a later thought. 

Consider now a set of problems $S$, and the past finding that there exists an optimal problem-specific reasoning length. Described in terms of reconstruction cycles, let $r_p$ reconstruction cycles be the optimal amount of reasoning for each problem $p\in S$. If the model were to always generate a reasoning chain containing at least $\overline{r}$ reconstruction cycles, where $\overline{r}$ is an upper bound of $\{r_p\}$, but the KV cache of only a small subset of these reconstruction cycles were kept, i.e., our described set $\overline{RC}_r$ after generating the $r^{\text{th}}$ reconstruction cycle, our initial belief was that by strategically choosing $\overline{RC}_r$ dynamically through time, the model could obtain and maintain the peak accuracy achievable through Budget Forcing. As detailed in the next section, our experimental results exceeded our expectations, with Min-Seek being able to not only obtain, but surpass the peak accuracy achievable through Budget Forcing, while also being robust to reasoning length.

Before explaining the details of choosing $\overline{RC}_r$, managing the KV cache efficiently for the potential generation of an unbounded number of reconstruction cycles will first be discussed. 

\subsection{KV Cache Management}

The modification made to the default DynamicCache \cite{hugging2025} of Hugging Face's Transformers package will now be described. The values are stored as usual, but the keys are saved without position embeddings. Before each new reconstruction cycle generation, a copy of the keys is made with contiguous position embeddings applied to them. The order of when reconstruction cycles were generated is preserved, but their precise position IDs are ignored, with the IDs $[0,|KV|-1]$ being encoded into the keys of $PT1$ and $\overline{RC}_r$ when generating the $r+1^{\text{th}}$ reconstruction cycle, where $|KV|$ is the current size of the $KV$ cache along the sequence length. 

Recall that the KV cache uses the update method \cite{hugging20252}

\vspace{5pt}

\noindent\texttt{K,V = cache.update(k\_t,v\_t,layer\_id)},

\vspace{5pt}

\noindent where the newly generated key and value at time step $t$ are inputted, and the full updated KV cache 
\texttt{(K,V)} for the current layer is returned. This is modified to

\vspace{5pt}

\noindent\texttt{K,V = cache.update(k\_t,k\_t\_no\_pos,v\_t, layer\_id)},

\vspace{5pt}

\noindent where \texttt{k\_t\_no\_pos} is the generated key without a position embedding. The tensors 
\texttt{(K,K\_no\_pos,V)}, containing the layer's keys, with and without position embeddings, and values are all updated, with only \texttt{(K,V)} being returned to compute the attention. After the generation of the current reconstruction cycle, only \texttt{(K\_no\_pos,V)} are returned to generate the next reconstruction cycle, with \texttt{K} being discarded. This method generated no measurable inference slow down in preliminary experiments. It does increase the required GPU memory, as two copies of keys are being stored, but as will be described, when using Min-Seek, only one reconstruction cycle, $|I_r|=1$, is all that is kept in the KV cache during reasoning, making our method overall memory efficient.    

Given that the position embeddings are now dynamically applied contiguously, as long as
the length of $PT_1$, each $RC_i$, and the final generated answer stay within some upper bound, say $u$, and the number of reconstruction cycles $|I_r|$ kept in the KV cache is bounded, $|I_r|\leq \overline{|I|}\in \NN$, such that the total upper bound on the size of the KV cache, $(\overline{|I|}+2)u$, is less than the maximum context length, then an unlimited number of reconstruction cycles can be generated, with an unbounded chain of reasoning outputted to the end user. Furthermore, given that the KV cache has a fixed upper bound, the method has linear computational complexity in the sequence length.

Besides allowing sequential test-time scaling beyond a model's maximum context length, we believe that dynamically applying position embeddings has further benefits, even in scenarios where the total token generation length fits within the maximum context length. RoPE \cite{su2024} is the most popular position embedding method, and what is used by the models in our experiments. A claimed benefit of using RoPE is decaying inter-token dependency with increasing relative distances, which implicitly assumes that it is desirable for tokens which are further apart to have less connection. Considering that $\overline{RC}_r$ may contain self-revisions generated at vastly different times during the iterative thought generation process, it is unclear why that information would be beneficial to encode, as they are all thoughts concerning the same problem. Self-revisions should be kept in the order that they were created, but we do not want past reconstruction cycles to slowly lose significance, simply because they are ``old". Even more important is the potential decrease in attention given to the prompt and the first generated thought, $PT_1$, which may be a contributing factor to the accuracy degradation which has been observed after generating many reconstruction cycles, with the model seemingly forgetting the original question. With our proposed KV cache, the current position ID is kept as small as possible by enforcing all position embeddings to be contiguous, which reduces any possible recency bias which could restrict the current reconstruction cycle from fully attending to $PT_1$ and $\overline{RC}_r$. 

\subsection{Selection of Reconstruction Cycles $\overline{RC}_r$}

Based on the past empirical finding that on average shorter answers are more accurate, our method is based on the simple rule of only keeping the shortest reconstruction cycles in the KV cache. From our empirical analysis, it was found to be sufficient to only keep the KV pairs of the currently shortest generated reconstruction cycle during reasoning to achieve stable and significantly higher accuracy compared to standard LRM generation. The motivation for this rule comes less so from trying to retain the most ``important" thoughts, but from trying to avoid logical pitfalls, where the model gets trapped in incorrect reasoning chains, which are on average longer, as the model struggles to come to an answer based on inconsistent or otherwise flawed reasoning.   

As a comparison, it is plausible that a reconstruction cycle $RC_r$ generated at step $r$ may be deemed insignificant when generated, but after further reasoning, at a future generation step $r+p$, in conjunction with subsequent thoughts, its importance may be realized, with the model wanting to attend to it. To accommodate this, assuming that important thoughts can be accurately detected, all past reconstruction cycles must be stored, and after every thought generation step, 
reconstruction cycles would need to be retrieved to include in the KV cache. In this setting, the contents of the KV cache at future states would in general depend on all of its past states. With our selection rule, based only on thought length, the KV cache already contains the KV pairs of the currently shortest reconstruction cycle, so its next state will only depend on its current state and the newly generated reconstruction cycle, by following the simple update rule,

\begin{alignat}{6}
V^l_{r}=\begin{cases}
	    V^l_{r-1}=[V^l_{PT_1}, V^l_{RC_m}] & \text{if }|RC_m|\leq |RC_{r}|,\nonumber\\
            [V^l_{PT_1}, V^l_{RC_r}] & \text{otherwise,}
		 \end{cases}
\end{alignat}

\noindent where $V^l_{r}$ are the values of layer $l$ to be used to generate $RC_{r+1}$, $V^l_{PT_1}$ are the values of $PT_1$, $V^l_{RC_m}$ are the values of the shortest (and oldest) reconstruction cycle up to $RC_{r-1}$, $m=\min(\argmin\limits_{i=1,...,r-1}|RC_i|)$, and $V^l_{RC_r}$ are the values of the currently generated $RC_{r}$. The keys without position embeddings are updated in the same way. We note that the KV cache is only updated when a strictly shorter reconstruction cycle has been generated, reducing the number of KV cache updates, improving generation speed, and we conjecture, the stability of our scaling method.


When a signal to generate an answer has been triggered (more details in the next section), we considered two variants of Min-Seek, which give as input to the model either 

1. ``Wait", or 

2. {\tt <\textbackslash think>}.


\noindent Inputting ``Wait" allows the model to complete one final reconstruction cycle before giving its answer, hence the answer produced by the model is based on the KV cache of the sequence $[PT_1,RC_m,RC_f]$, where $RC_f$ is the final reconstruction cycle. Inputting {\tt <\textbackslash think>} forces the model to directly give an answer, which is generated from the KV cache of the tokens $[PT_1,RC_m]$. As will be seen in the next section, where two LRMs were tested, it was found that for the smaller of the two models, it was sufficient to use variant 2, whereas for the larger tested model, we were able to leverage its greater reasoning capacity by using variant 1.

At first glance, Min-seek, consisting of essentially one extra thought, may seem too simple to be able to achieve significantly improved accuracy, but it is worth noting that each new reconstruction cycle can build off of and contain essential information derived from past reconstruction cycles. Following a recursive pattern, where the $n^{\text{th}}$ previous minimum length reconstruction cycle is denoted as $RC_{m[-n]}$, and considering their KV caches, $RC_m$ is a function of $[PT_1,RC_{m[-1]}]$, $RC_{m[-1]}$ is a function of $[PT_1,RC_{m[-2]}]$, and so on. A small example of using Min-Seek with and without our proposed custom KV cache is given in Figure \ref{fig:experiments}. 

\section{Experiments}
\label{experiments}

Our proposed sequential test-time scaling method, Min-Seek, was tested against the most comparable test-time scaling method for LRMs, Budget Forcing \cite{muennighoff2025,zeng2025}. 

\subsection{Models \& Datasets}

The experiments were conducted using two reasoning models distilled from DeepSeek-R1: 
DeepSeek-R1-Distill-Qwen-1.5B \cite{DS1B} and DeepSeek-R1-Distill-Qwen-7B \cite{DS7B}, with five challenging reasoning datasets:

\noindent 1) AIME 2024 \cite{H4AIME24}: 2024 AIME I and II examinations with three digit non-negative integer answers.

\noindent 2) AMC \cite{PNAMC}: AMC12 2022 and 2023 examinations, where the originally 4-choice questions were modified to have integer answers. 

\noindent 3) GPQA Diamond \cite{reindata2024}: ``Google-proof" 4-choice questions in biology, physics, and chemistry that experts are able to answer correctly, but the majority of non-experts cannot, even with unrestricted use of the internet \cite[Section 2]{rein2024}. 

\noindent 4) MATH-500 \cite{H4MATH500}: The test-set problems from the original MATH dataset \cite{hendrycks2021} used in \cite[Appendix C]{lightman2024} for evaluation, formatted using \LaTeX\ to express mathematical quantities.

\noindent 5) MMLU-Pro \cite{MMLUPro}: Evaluation set of 10-choice questions spanning 14 disciplines from biology to psychology \cite[Figure 3(a)]{wang2024}.

\subsection{Implementation Details}

For both tested methods, reconstruction cycles were repeatedly generated until one of three conditions were met:

1. Generation token limit exceeded: We set a (soft) limit of $2^{15}$ generation tokens. Once this quantity is exceeded, no further reconstruction cycles are generated, and the model is forced to generate a final answer.  

2. Reconstruction cycle limit reached: The majority of experiments were done with a limit on the total number of reconstruction cycles. As our work is interested in long, potentially unbounded reasoning, limiting the number of reconstruction cycles was done to verify the stability of Min-Seek through time, and to show its strong performance compared to Budget Forcing, even if the number of generated reconstruction cycles were to be tuned for accuracy.

3. No generated {\tt <\textbackslash think>} token: Even though the LRMs were instructed to use special tokens to enclose their reasoning and final answer, their generation did not always follow this pattern, which was deemed to be out of our control. For instances where no {\tt <\textbackslash think>} token was generated, test-time scaling was not continued, with the generated output being considered the final reconstruction cycle and answer for Budget Forcing and variant 1 of Min-Seek. For variant 2, this generation was discarded, and the previously inputted ``Wait" was

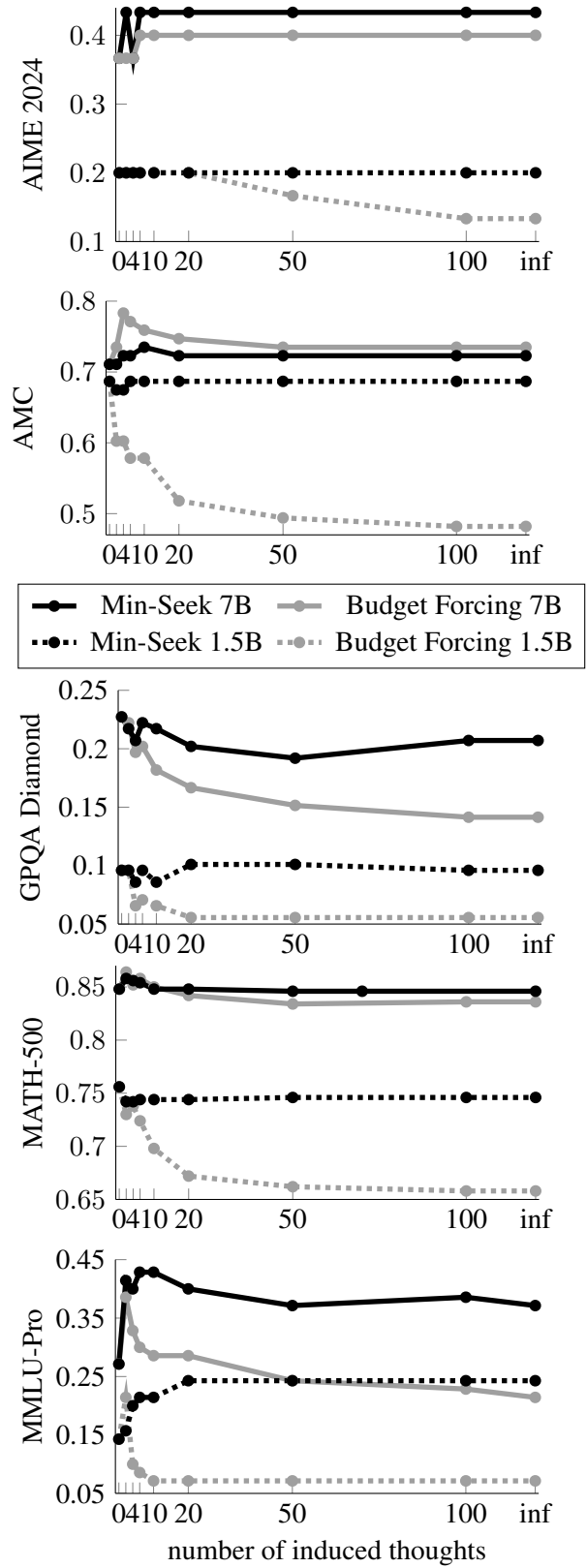
\begin{figure}[H]
	\centering
	\pgfplotsset{width=5.75cm,height=3.2cm,compat=1.3}
	\begin{tikzpicture}
		\pgfplotsset{scale only axis}
		\begin{axis}[
			axis y line*=left,axis x line*=bottom,
			xmin=-1,xmax=121,ymin=0.10,ymax=0.44,
			xtick=data,ytick={0.1,0.2,...,0.4},
			xticklabels={0,,4,,10,20,50,100,inf},
			ylabel=AIME 2024,
			y label style={at={(axis description cs:-0.15,.5)},anchor=south}]
			\addplot[draw=black,line width=2pt,mark=*,mark size=1.25pt,mark options={fill=black}]
			table[x=x,y=y]{min_thought_7B_aime.dat};\label{plot_2}
			\addplot[draw=lightgray!50!gray,line width=2pt,mark=*,mark size=1.25pt,mark options={fill=lightgray!50!gray}]
			table[x=x,y=y]{add_thought_7B_aime.dat};\label{plot_1}	
			\addplot[draw=lightgray!50!gray,dotted,line width=2pt,mark=*,mark size=1.25pt,mark options={solid,fill=lightgray!50!gray}]
			table[x=x,y=y]{add_thought_1B_aime.dat};\label{plot_11}	
			\addplot[draw=black,dotted,line width=2pt,mark=*,mark size=1.25pt,mark options={solid,fill=black}]
			table[x=x,y=y]{min_thought_1B_aime.dat};\label{plot_22}
		\end{axis}
	\end{tikzpicture}
	\begin{tikzpicture}
		\pgfplotsset{scale only axis}
		\begin{axis}[
			axis y line*=left,axis x line*=bottom,
			xmin=-1,xmax=121,ymin=0.47,ymax=0.8,
			xtick=data,ytick={0.5,0.6,...,0.8},
			xticklabels={0,,4,,10,20,50,100,inf},
			ylabel=AMC,
			y label style={at={(axis description cs:-0.15,.5)},anchor=south}]
			\addplot[draw=lightgray!50!gray,line width=2pt,mark=*,mark size=1.25pt,mark options={fill=lightgray!50!gray}]
			table[x=x,y=y]{add_thought_7B_amc.dat};\label{plot_3}	
			\addplot[draw=black,line width=2pt,mark=*,mark size=1.25pt,mark options={fill=black}]
			table[x=x,y=y]{min_thought_7B_amc.dat};\label{plot_4}
			\addplot[draw=lightgray!50!gray,dotted,line width=2pt,mark=*,mark size=1.25pt,mark options={solid,fill=lightgray!50!gray}]
			table[x=x,y=y]{add_thought_1B_amc.dat};\label{plot_33}	
			\addplot[draw=black,dotted,line width=2pt,mark=*,mark size=1.25pt,mark options={solid,fill=black}]
			table[x=x,y=y]{min_thought_1B_amc.dat};\label{plot_44}
		\end{axis}
		\matrix[matrix of nodes,anchor=west,xshift=-1.2cm,yshift=-1.25cm,inner sep=0.2em,draw]
		{\ref{plot_2}&Min-Seek 7B&\ref{plot_1}&Budget Forcing 7B\\
			\ref{plot_22}&Min-Seek 1.5B&\ref{plot_11}&Budget Forcing 1.5B\\};
	\end{tikzpicture}
	\begin{tikzpicture}
		\pgfplotsset{scale only axis}
		\begin{axis}[
			axis y line*=left,axis x line*=bottom,
			xmin=-1,xmax=121,ymin=0.05,ymax=0.25,
			xtick=data,ytick={0.05,0.1,...,0.25},
			xticklabels={0,,4,,10,20,50,100,inf},
			yticklabels={0.05,0.1,0.15,0.2,0.25},
			ylabel=GPQA Diamond,
			y label style={at={(axis description cs:-0.15,.5)},anchor=south}]
			\addplot[draw=lightgray!50!gray,line width=2pt,mark=*,mark size=1.25pt,mark options={fill=lightgray!50!gray}]
			table[x=x,y=y]{add_thought_7B_gpqa.dat};\label{plot_5}
			\addplot[draw=black,line width=2pt,mark=*,mark size=1.25pt,mark options={fill=black}]
			table[x=x,y=y]{min_thought_7B_gpqa.dat};\label{plot_6}
			\addplot[draw=lightgray!50!gray,dotted,line width=2pt,mark=*,mark size=1.25pt,mark options={solid,fill=lightgray!50!gray}]
			table[x=x,y=y]{add_thought_1B_gpqa.dat};\label{plot_55}	
			\addplot[draw=black,dotted,line width=2pt,mark=*,mark size=1.25pt,mark options={solid,fill=black}]
			table[x=x,y=y]{min_thought_1B_gpqa.dat};\label{plot_66}
		\end{axis}
	\end{tikzpicture}
	\begin{tikzpicture}
		\pgfplotsset{scale only axis}
		\begin{axis}[
			axis y line*=left,axis x line*=bottom,
			xmin=-1,xmax=121,ymin=0.65,ymax=0.87,
			xtick=data,ytick={0.65,0.7,...,0.85},
			xticklabels={0,,4,,10,20,50,100,inf},
			ylabel=MATH-500,
			y label style={at={(axis description cs:-0.15,.5)},anchor=south}]
			\addplot[draw=lightgray!50!gray,line width=2pt,mark=*,mark size=1.25pt,mark options={fill=lightgray!50!gray}]
			table[x=x,y=y]{add_thought_7B_math500.dat};\label{plot_7}	
			\addplot[draw=black,line width=2pt,mark=*,mark size=1.25pt,mark options={fill=black}]
			table[x=x,y=y]{min_thought_7B_math500.dat};\label{plot_8}
			\addplot[draw=lightgray!50!gray,dotted,line width=2pt,mark=*,mark size=1.25pt,mark options={solid,fill=lightgray!50!gray}]
			table[x=x,y=y]{add_thought_1B_math500.dat};\label{plot_77}	
			\addplot[draw=black,dotted,line width=2pt,mark=*,mark size=1.25pt,mark options={solid,fill=black}]
			table[x=x,y=y]{min_thought_1B_math500.dat};\label{plot_88}
		\end{axis}
	\end{tikzpicture}
	\begin{tikzpicture}
		\pgfplotsset{scale only axis}
		\begin{axis}[
			axis y line*=left,axis x line*=bottom,
			xmin=-1,xmax=121,ymin=0.05,ymax=0.45,
			xtick=data,ytick={0.05,0.15,...,0.45},
			xticklabels={0,,4,,10,20,50,100,inf},
			xlabel=number of induced thoughts,
			ylabel=MMLU-Pro,
			yticklabels={0.05,0.15,0.25,0.35,0.45},
			y label style={at={(axis description cs:-0.15,.5)},anchor=south},
			x label style={at={(axis description cs:.5,-0.35)},anchor=south}]
			\addplot[draw=lightgray!50!gray,line width=2pt,mark=*,mark size=1.25pt,mark options={fill=lightgray!50!gray}]
			table[x=x,y=y]{add_thought_7B_mmlupro.dat};\label{plot_9}	
			\addplot[draw=black,line width=2pt,mark=*,mark size=1.25pt,mark options={fill=black}]
			table[x=x,y=y]{min_thought_7B_mmlupro.dat};\label{plot_10}
			\addplot[draw=lightgray!50!gray,dotted,line width=2pt,mark=*,mark size=1.25pt,mark options={solid,fill=lightgray!50!gray}]
			table[x=x,y=y]{add_thought_1B_mmlupro.dat};\label{plot_99}	
			\addplot[draw=black,dotted,line width=2pt,mark=*,mark size=1.25pt,mark options={solid,fill=black}]
			table[x=x,y=y]{min_thought_1B_mmlupro.dat};\label{plot_100}
		\end{axis}
	\end{tikzpicture}
	\caption{Average accuracy, from top to bottom, for AIME 2024, AMC, GPQA Diamond, MATH-500, and MMLU-Pro, for an increasing number of induced thoughts (reconstruction cycles) using our proposed method Min-Seek, and Budget Forcing, for DeepSeek-R1-Distill-Qwen-7B (7B) and DeepSeek-R1-Distill-Qwen-1.5B (1.5B).} \label{fig3}
\end{figure}

\noindent replaced by {\tt <\textbackslash think>} to directly generate the final answer based on the KV cache of $[PT_1,RC_m]$. This was viewed as the fairest approach, as incorrectly trying to parse the generated reasoning and answer tokens may result in unnatural token generation and a non-representative loss in accuracy.

Following \cite{muennighoff2025}, for each dataset and model, Min-Seek and Budget Forcing were tested by inducing at most $M=0$ (standard model generation), $2$, $4$, and $6$ reconstruction cycles. This was extended to generating at most $M=10$, $20$, $50$, $100$, and an unbounded number of reconstruction cycles, referred to as inf(inite) in the plots, in order to fully test the reasoning stability and improved accuracy of Min-Seek without any explicit limit on its number of reconstruction cycles. A single generation was performed for each dataset sample and choice of $M$ using the same random seed for all models and methods. The default LRM generation configuration was used, including a temperature of $0.6$ and a top\_p value of $0.95$. 

\subsection{Results}

For each model, task, and test-time scaling method, the average accuracy when using up to $M\in\{0,2,4,6,10,20,50,100,\text{inf}\}$ reconstruction cycles was divided by the average accuracy when using $M=0$, resulting in standard LRM generation having a normalized accuracy of 1, and the normalized accuracy of each method with $M>0$ indicating its percentage improvement over standard generation. For each model, method, and value of $M$, the average normalized accuracies over all reasoning tasks are plotted in Figure \ref{fig1}, with the raw accuracies for each task plotted in Figure \ref{fig3}.  

Examining the overall average accuracy of Min-Seek displayed in Figure \ref{fig1}, it can be seen that it is able to stabilize sequential test-time scaling, and generate significantly improved accuracy compared to standard LRM generation. For DeepSeek-R1-Distill-Qwen-1.5B, Budget Forcing is able to achieve a peak normalized accuracy of $1.069$ with $M=2$, whereas Min-Seek is able to outperform Budget Forcing with $M=2$ for all $M\geq 6$, maintaining an accuracy in the range of $[1.076,1.148]$, and stabilizing to an accuracy of $1.137$ for $M\geq 100$. For DeepSeek-R1-Distill-Qwen-7B, Budget Forcing achieves a peak accuracy of $1.090$ at again $M=2$, with Min-Seek dominating Budget Forcing, maintaining an improvement in average normalized accuracy of at least $0.045$ for all choices of $M>0$, and stabilizing to an accuracy $>1.095$ for $M\geq 100$. Examining the per-task accuracies in Figure \ref{fig3}, with the exception of AMC using DeepSeek-R1-Distill-Qwen-7B, Min-Seek is largely dominating Budget Forcing for all $M>0$.

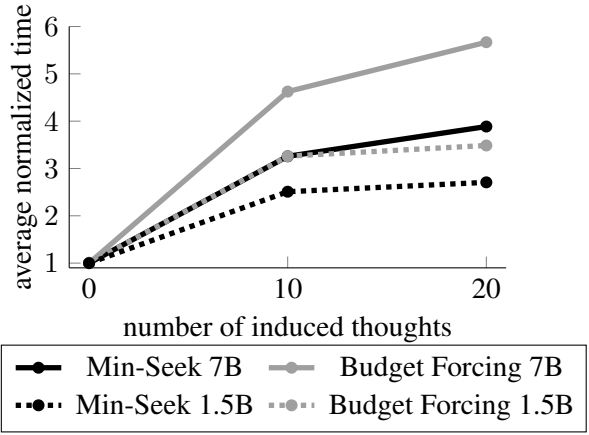
\begin{figure}[t]
\centering
\pgfplotsset{width=5.75cm,height=3.2cm,compat=1.3}
\begin{tikzpicture}
\pgfplotsset{scale only axis}
\begin{axis}[
axis y line*=left,axis x line*=bottom,
xmin=-1,xmax=21,ymin=0.9,ymax=6,
xtick=data,ytick={1,2,...,6},
xticklabels={0,10,20},
ylabel=average normalized time,
xlabel=number of induced thoughts,
y label style={at={(axis description cs:-0.05,.5)},anchor=south},
x label style={at={(axis description cs:.5,-0.35)},anchor=south}]
\addplot[draw=black,line width=2pt,mark=*,mark size=1.25pt,mark options={fill=black}]
table[x=x,y=y]{min_time_7B_norm.dat};\label{plot_101}
\addplot[draw=lightgray!50!gray,line width=2pt,mark=*,mark size=1.25pt,mark options={fill=lightgray!50!gray}]
table[x=x,y=y]{add_time_7B_norm.dat};\label{plot_102}	
\addplot[draw=lightgray!50!gray,dotted,line width=2pt,mark=*,mark size=1.25pt,mark options={solid,fill=lightgray!50!gray}]
table[x=x,y=y]{add_time_1B_norm.dat};\label{plot_201}	
\addplot[draw=black,dotted,line width=2pt,mark=*,mark size=1.25pt,mark options={solid,fill=black}]
table[x=x,y=y]{min_time_1B_norm.dat};\label{plot_202}
\end{axis}
\matrix[matrix of nodes,anchor=west,xshift=-0.9cm,yshift=-1.6cm,inner sep=0.2em,draw]
{\ref{plot_101}&Min-Seek 7B&\ref{plot_102}&Budget Forcing 7B\\
\ref{plot_202}&Min-Seek 1.5B&\ref{plot_201}&Budget Forcing 1.5B\\};
\end{tikzpicture}
\caption{Average normalized computation time over 5 reasoning tasks. A maximum of 0, 10, and 20 thoughts (reconstruction cycles) were induced using our proposed method Min-Seek, and Budget Forcing, for DeepSeek-R1-Distill-Qwen-7B (7B) and DeepSeek-R1-Distill-Qwen-1.5B (1.5B).} \label{fig4}
\end{figure}

In order to accurately measure computation time, the experiments using $M\in\{0,10,20\}$ were done in isolation, with their average normalized computation time plotted in Figure \ref{fig4}. The generation speed up from using Min-Seek can be observed, with Budget Forcing being on average $44.0\%$ and $29.4\%$ slower than Min-Seek for DeepSeek-R1-Distill-Qwen-7B and 1.5B, respectively, demonstrating Min-Seek's improved efficiency by keeping only one reconstruction cycle in the KV cache during reasoning.

\section{Conclusion}
\label{conclusion}

This work presented Min-Seek, a sequential test-time scaling method which corrects the instability and accuracy degradation from generating long reasoning chains, achieving a significant improvement in accuracy compared to standard model generation, as well as Budget Forcing sequential test-time scaling, for a wide range of generated thoughts. Min-Seek is simple, requiring only the retention of a single induced thought during reasoning, resulting in a significant speed up in generation time compared to Budget Forcing. With a proposed custom KV cache which stores keys without position embeddings, a model's maximum context length can be bypassed, allowing for an unbounded number of thoughts to be generated with linear computational complexity. 

\section*{Limitations}

This work proposes a training-free method which can only be used with a pre-existing LRM. For some tasks, the standard generation of an LRM may be sufficient to correctly respond to a given prompt, with Min-Seek only increasing the computation time. Min-Seek is most applicable for challenging reasoning tasks which the model struggles with, and in particular for when the user wants to verify the reasoning and observe the solution approaches considered by the model before arriving at its solution.  

\bibliography{custom.bib}

\end{document}